\documentclass[11pt]{article}
\usepackage{eacl2017}
\usepackage{times}
\usepackage{url}
\usepackage{latexsym}
\usepackage{graphicx}

\eaclfinalcopy %

\title{Using Fisher's Exact Test to Evaluate Association Measures for N-grams}

\author{Yves Bestgen \\
  Laboratoire d'analyse statistique des textes - LAST \\
  Institut de recherche en sciences psychologiques \\
  Universit\'e catholique de Louvain \\
  Place Cardinal Mercier, 10 1348 Louvain-la-Neuve, Belgium \\
  \texttt{yves.bestgen@uclouvain.be} \\}

\date{}

\begin{document}
\maketitle
\begin{abstract}
To determine whether some often-used lexical association measures assign high scores to n-grams that chance could have produced as frequently as observed, we used an extension of Fisher's exact test to sequences longer than two words to analyse a corpus of four million words. The results, based on the precision-recall curve and a new index called chance-corrected average precision, show that, as expected, simple-ll is extremely effective. They also show, however, that MI3 is more efficient than the other hypothesis tests-based measures and even reaches a performance level almost equal to simple-ll for 3-grams. It is additionally observed that some measures are more efficient for 3-grams than for 2-grams, while others stagnate.
\end{abstract}
\section{Introduction}
Evaluating the effectiveness of lexical association measures (AMs) to identify multiword expressions (MWE) has caught the attention of many researchers for over twenty years (e.g., Church and Hanks, 1990; Evert and Krenn, 2005; Ramish, 2015). In the case of 2-grams, a large number of measures have been compared and several syntheses are available (e.g., Evert, 2009; Pecina, 2010). Measures of association for longer n-grams have received much less attention, leading Evert (2009) to place such measures on the field's to-do list. A similar conclusion was reached by Gries (2010) and Nerima et al. (2010), among others.

This situation has probably led some researchers to focus on raw co-occurrence frequency as the criterion for selecting various type of MWE (e.g., Biber 2009; Frantzi et al., 2000; O'Keeffe et al., 2007; Wermter and Hahn, 2006). In the case of 2-grams, however, it is widely recognized that frequency alone is insufficient to distinguish genuine sequences from sequences that chance could have produced as frequently as observed in the corpus (Evert, 2009; Gries, 2010). According to Ellis et al. (2015), this same phenomenon should also be observed with longer sequences. Yet, the longer a sequence is, the less likely it is to occur by chance, even when it is composed of frequent words. The question is thus how long a sequence must be to ensure that raw frequency selects only sequences that are unlikely to result from chance. 

This question is indeed far more general. It can be asked about almost any AMs since they mainly aim at highlighting sequences whose association score is sufficiently greater than that expected by chance. If this is obvious for the AMs based on hypothesis testing, such as \textit{z} or \textit{simple-ll}, it is also true of effect size measures, such as \textit{MI} and its variants, because these measures also compare the observed frequencies to the expected frequencies, even if they do not derive a test statistic. Comparing frequency to AMs is all the more important because Biber (2009) has made quite strong claims against the use of AMs such as MI to select the most interesting n-grams in a corpus.

This paper proposes to evaluate this better-than-chance assumption in the case of sequences longer than two words by means of the Fisher's exact test, the inferential test considered to be the most sensible and accurate statistical test for analysing the frequency of 2-grams in a corpus (Evert, 2009; Moore, 2004; Pedersen, 1996; Stefanowitsch and Gries, 2003). It is important to note that this study only aims to develop a complementary point of view on the effectiveness of some AMs. In no case can it replace an assessment based on a gold standard list established manually or on the benefits that automatically-extracted MWE can provide to NLP applications.
The next section briefly describes the extension of Fisher's exact test to sequences of more than two words. We then present an experiment conducted on a corpus of four million words to analyse the performance of six AMs for extracting 2-grams to 4-grams.
\section{Applying Fisher's Exact Test to 3-grams and Longer N-grams}
As mentioned above, Fisher's exact test can be used to calculate the probability that chance alone could have produced at least as many instances of a 2-gram as the number actually observed in the corpus (Evert, 2009; Jones and Sinclair 1974; Pedersen et al., 1996). However, its use for analysing longer sequences is problematic because these sequences require the construction of contingency tables of three or more dimensions. As exact tests proposed for such tables (Zelterman et al., 1995) are not suited to the study of word sequences, due to the sample size and the large number of tests to be performed, Bestgen (2014, 2018) proposed the application to 3-grams and longer n-grams of the usual procedure to estimate the Fisher's exact probability when the use of the hypergeometric formula is not possible (Agresti, 1992; Pedersen, 1996). This method rests on a Monte Carlo permutation procedure to generate a random sample of the possible contingency tables, given the marginal totals, the needed probability corresponding to the proportion of these tables that give rise to a value at least as extreme as that observed. Given that a corpus is a long sequence of graphic forms, its permutation is simply the random mixing of these forms. Thus, any permutation of the order of the tokens in the corpus generates a random contingency table for every original 2-grams, solving the problem of the large number of tests to be performed. If this procedure is iterated, the proportion of permutations in which the 2-gram is at least as frequent as in the original corpus is an estimate of the exact probability. This estimation procedure can be straightforwardly generalized to longer n-grams by counting, in each random permutation, the 3-grams, 4-grams, etc. 

Bestgen (2014) showed that this procedure allows an almost perfect estimation of the Fisher's exact probability for all the 2-grams in a corpus, with the caveat that the number of permutations carried out limits the precision of the probability. The procedure's major weakness is thus its computational cost. Therefore, it does not lead to a viable AM in itself, but it might help in the assessment of proposed AMs.

\section{Experiment}
This experiment aims at evaluating, by means of Fisher's exact test, the effectiveness of a series of AMs that have been proposed to extract sequences of two to four words.
\subsection{Materials and Methods}
{\bf Corpus}: The analyses were conducted on a corpus of 5,529,378 tokens, including 4,232,259 words of spontaneous conversations from the demographic spoken section of the British National Corpus. This corpus, similar in content to the corpora used in several studies on lexical bundles (Biber, 2009), was chosen because it was thought to be of sufficient size to extract MWE while not being so large that a sufficient number of permutations could not be performed.
\newline {\bf Corpus Processing}: The version of the corpus POS-tagged by CLAWS was employed. All orthographic forms (words, but also numbers, symbols and punctuation) detected by CLAWS were considered as tokens to be randomly mixed. Twenty million permutations were performed. To obtain the sequences and their frequency in the original corpus and in each permutation, only strings of uninterrupted (and lowercased) word-forms were taken into account. Thus, any punctuation mark or sequence of characters that did not correspond to a word interrupted the sequence extraction.
Because association measures computed for very rare n-grams are known to be unreliable (Evert, 2009), we only analysed sequences that occurred at least three times in the original corpus. 
\newline {\bf Selected Lexical Association Measures}: For these analyses, a series of simple association measures were selected based on their popularity in studies of 2-grams and longer sequences: \textit{MI} and \textit{t}-score (\textit{t}; Church et al., 1991), \textit{z} (Berry-Rogghe, 1973), \textit{simple-ll} (Evert, 2009) and the raw frequency (\textit{f}). We also analysed \textit{MI3}, a heuristic modification of \textit{MI}, proposed to reduce its tendency to assign inflated scores to rare words that occur together (Daille, 1994). However, as \textit{MI3} violates one convention for AMs by assigning, in some cases, positive scores to n-grams that co-occur less frequently than expected by chance (Evert, 2009, p. 1226), we modified the formula in such a way that these n-grams were assigned a negative score and, therefore, appear at end of the list\footnote{Several modifications were attempted, but since they led to the same performance, we decided to simply go back to the usual MI formula for any n-gram that co-occurs less often than expected by chance. Other transformations could, however, be better for specific usage.}. A similar problem arises for \textit{simple-ll} and, in this case, we used the signed version recommended by Evert (2009). The formulas of these indices for the 3-grams and 4-grams are identical to those used for the 2-grams, the expected frequency being calculated using the usual extension of the 2-gram formula (i.e., by multiplying the product of the marginal probabilities of all words by the corpus size; but see discussion in Gries (2010)).
\subsection{Analyses and Results}
The main research question this study attempts to answer is whether some often-used AMs assign high scores to n-grams whose frequencies are likely to result from chance according to an inferential test. The rejection level used was set to the classic value of 0.05. Holm's sequential procedure (Gries, 2005; Holm, 1979) was used to take into account the large number of tests, which increases the probability of erroneously deciding that at least one test is statistically significant. This procedure ensures a family-wise error rate of 0.05 for each n-gram length, thus allowing for their comparison.

Since Fisher's exact test acts in the analyses as a gold standard list, the results were summarized in the usual way by the precision-recall (PR) curves and the average precision (AP). As a baseline, we used the expected AP of a system that randomly ranks the n-grams, which is, according to Biemann et al. (2013), is equal to the proportion of significant n-grams. The number of tested n-grams (i.e., those with a frequency of at least three), the number of significant n-grams and the baseline AP for the three n-gram lengths are given in Table 1. It can be seen that almost all the 4-grams are considered by the inferential test to be statistically significant, but also that the baseline is much higher for the 3-grams than for the 2-grams.

\begin{table}
\small
\centering
\begin{tabular}{|l|rrr|}
\hline
 & \bf  2-gram & \bf  3-gram & \bf  4-gram \\ \hline
\#Types & 96881 & 111164 & 53334 \\
\#Sig & 27063 & 70942 & 52825 \\
Baseline AP & 0.279 & 0.638 & 0.990 \\
\hline
\end{tabular}
\caption{Details of the three n-gram lengths}
\end{table}
These findings have two implications for the analyses reported below. Firstly, they were conducted only on the 2-grams and 3-grams. For the 4-grams, the Fisher's criterion was irrelevant since almost all 4-grams present in the corpus at least three times passed the inferential test successfully.
\begin{figure*}
\centering
\begin{minipage}{.5\textwidth}
  \centering
  \includegraphics[width=.81\linewidth]{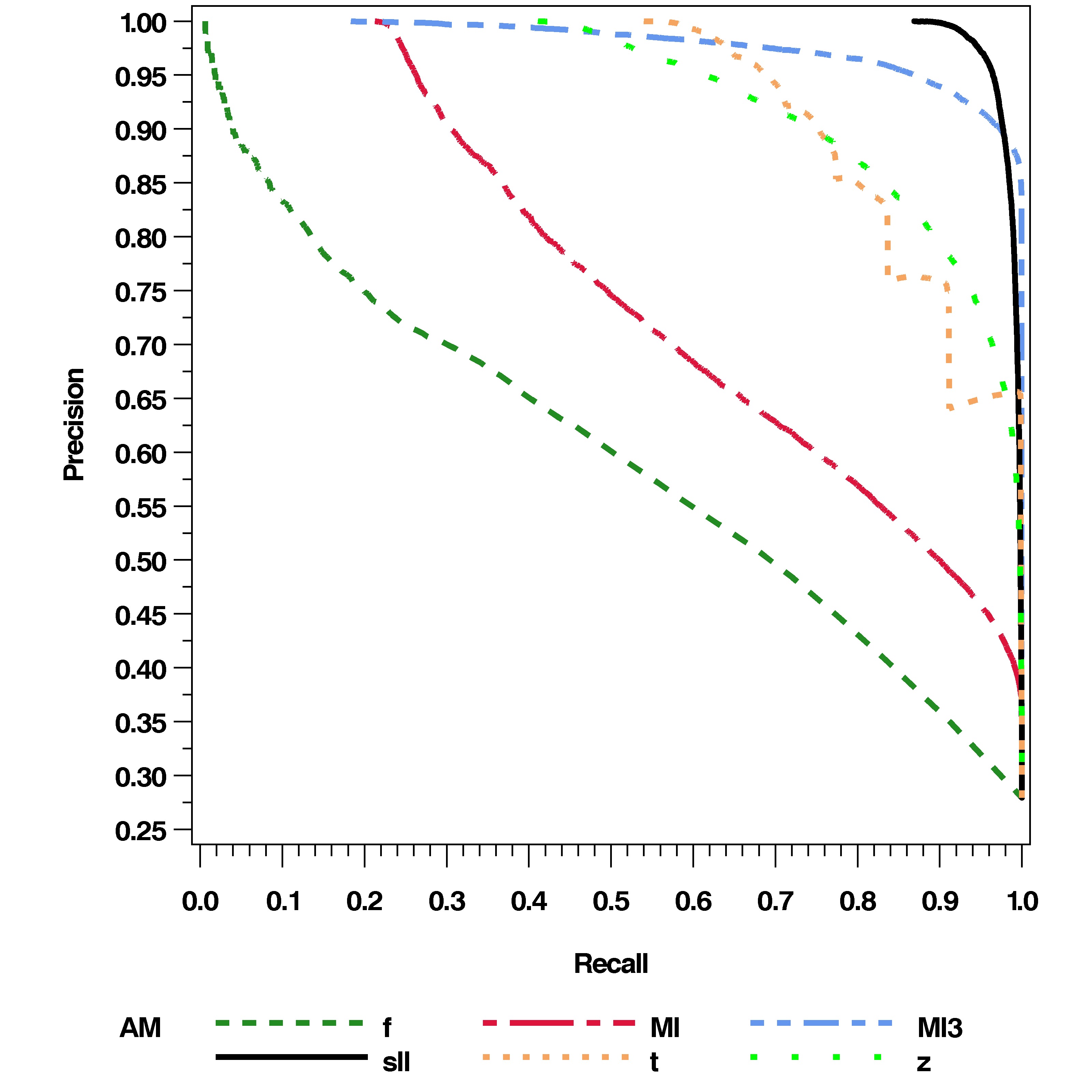}
  \caption{PR curves for the 2-grams}
\end{minipage}%
\begin{minipage}{.5\textwidth}
  \centering
  \includegraphics[width=.81\linewidth]{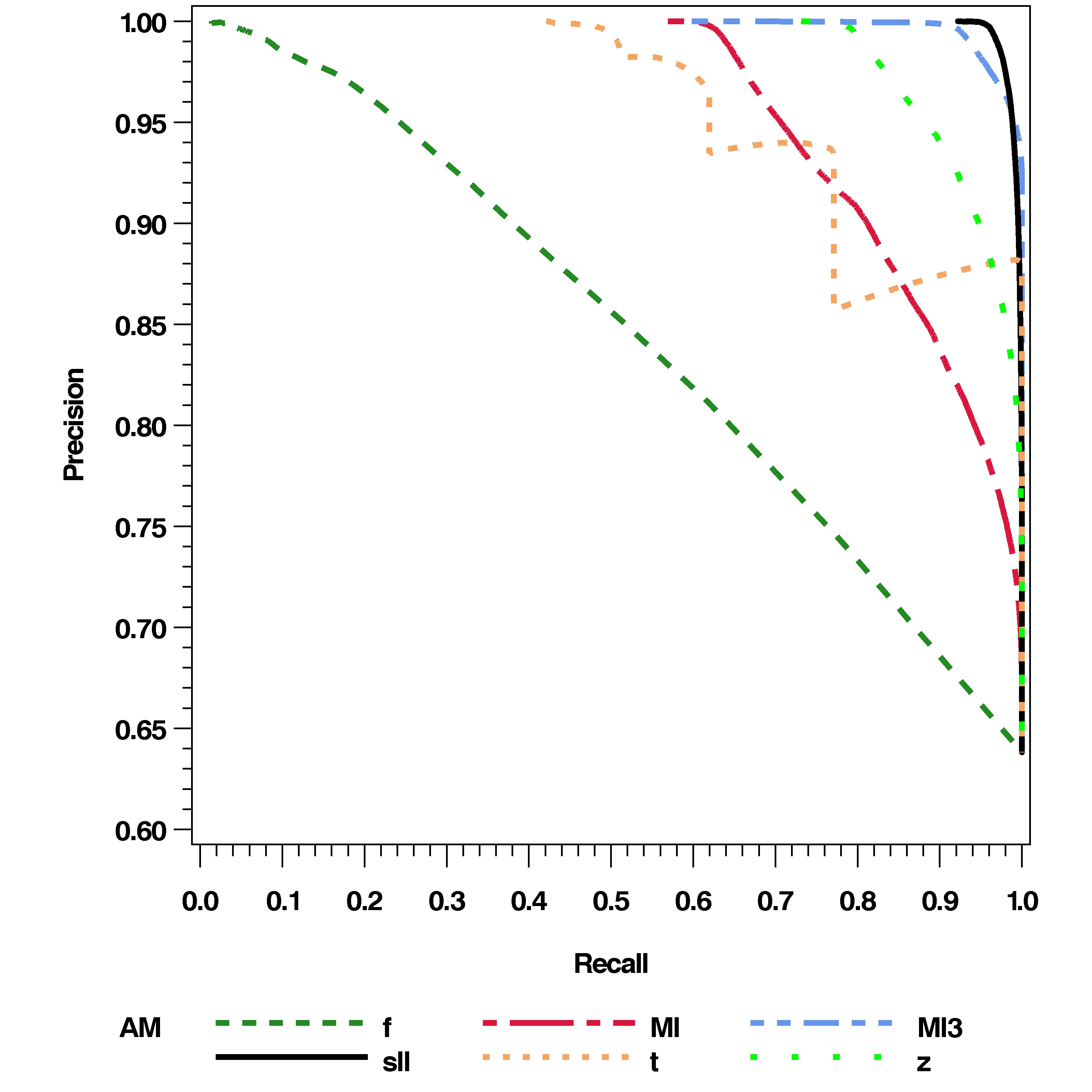}
  \caption{PR curves for the 3-grams}
\end{minipage}
\end{figure*}

The second consequence concerns the comparison of the performances of the AMs for the 2-grams and the 3-grams, as better performance should be expected in the second case. For this reason, in addition to the standard AP, we report a new index called chance-corrected AP obtained by inserting into the usual Kappa formula (Cohen, 1960; Powers, 2012) the baseline AP, since it is the level of performance that should, on average, produce a random ordering of the n-grams.

Figures 1 and 2 show the PR curves for the 2-grams and 3-grams. For each AM, the line begins at the first n-gram that does not pass the inferential test. Up to this point, therefore, the precision is 100\%. To facilitate the comparison of the figures, the starting point on the y-axis was set to a value just below the baseline AP. Table 2 reports the AP and the chance-corrected AP.

Looking first at the PR curve for the 2-grams, the results are largely in line with what one would expect. \textit{Simple-ll}, which is the AM that produces the ranking most similar to Fisher's exact test (Evert, 2009), is by far the best AM, reaching an almost perfect level of performance. It is followed by \textit{MI3}, \textit{z} and \textit{t}\footnote{The distinctive shape of the PR curve for \textit{t} is due to its formula, which assigns almost identical scores to the series of n-grams that occur in the corpus at the same low frequency and have an expected frequency close to 0. As these n-grams are considered as significantly too frequent by the inferential test, they form a long series of slightly increasing PR scores.}, which are, respectively, a version of \textit{MI}, which aims at the reduction of the impact of small frequencies, and two other inferential tests. The raw frequency is the least accurate, as expected, its PR curve showing that it ranks at the top of the list many 2-grams that chance alone could have produced as frequently as observed.

For the 3-grams, the difference between \textit{simple-ll} and \textit{MI3} is extremely small, as confirmed by the AP (0.9983 vs. 0.9977). This result was unexpected. \textit{Z} fares slightly worse. \textit{MI} is almost as effective as \textit{t}, another unexpected result, though the frequency threshold must have helped \textit{MI}. Raw frequency again brings up the rear with a significantly worse performance.
\begin{table}
\small
\centering
\begin{tabular}{|l|cc|cc|}
\hline
& \multicolumn{2}{c|}{AP} &  \multicolumn{2}{c|}{CcAP} \\ 
AM & 2-gram & 3-gram & 2-gram & 3-gram \\ \hline
f & 0.588 & 0.824 & 0.429 & 0.515 \\
MI & 0.756 & 0.957 & 0.662 & 0.881 \\
MI3 & 0.947 & 0.998 & 0.926 & 0.994 \\
z & 0.934 & 0.985 & 0.909 & 0.959 \\
t & 0.931 & 0.959 & 0.905 & 0.886 \\
simple-ll & 0.993 & 0.998 & 0.990 & 0.995 \\
\hline
\end{tabular}
\caption{AP and chance-corrected AP (CcAP) }
\end{table}

The chance-corrected APs for the 2-grams and 3-grams show that \textit{t} does not improve at all while \textit{MI}, \textit{MI3} and \textit{z} improve sharply.

One objective of these analyses was to determine whether the AMs give priority in the ranked list to n-grams that chance could have produced as frequently as observed. It is thus interesting to notice the position in the ranking of the first n-gram rejected by the inferential test, even if one must be cautious with such a punctual observation. It corresponds in the graphics to the leftmost point of each PR curve. It is provided in terms of recall and must, therefore, be interpreted with reference to the number of significant n-grams. For \textit{simple-ll}, the first non-significant 2-gram is observed at a recall of 87\%. It thus occurs after the 23,000th 2-gram in the list. For the 3-grams, it is beyond the 65,000th position. These values are clearly better than those obtained by the other AMs, suggesting it might be useful to filter the other AM ranked lists on this basis.

\section{Conclusion}
The objective of this study was to provide a new perspective on the effectiveness of lexical AMs for extracting MWE from corpora. The analyses demonstrate very large differences between the various AMs based on hypothesis testing. The near-perfect performance of \textit{simple-ll} extends to 3-grams Evert's conclusion (2009) that \textit{simple-ll} is a good approximation of Fisher's exact test for 2-grams. \textit{Z} is less effective. The \textit{t}-score is even much less so, and does not improve from 2-grams to 3-grams. Regarding measures of effect size, \textit{MI3}\footnote{The analyses were also performed using the uncorrected formula for \textit{MI3}. They indicated that the correction improved the performance for the 2-grams only.} is significantly more efficient than \textit{MI} and, for the 3-grams, reached a performance roughly equivalent to \textit{simple-ll}. It certainly deserves to be further evaluated in depth.

The criterion used (i.e., does an AM assign high scores to n-grams that chance could have produced as frequently as observed) suggests that it might be worthwhile to use \textit{simple-ll} to filter the rankings generated by another AM. Before recommending such a procedure, however, it is necessary to ensure that it does indeed improve the performance of the AMs in some NLP applications.

A potentially important limitation of this study is that it was conducted on only one corpus, composed exclusively of conversations and of a limited size. If analysing a larger corpus would greatly increase the time required to obtain the permutations, it could be interesting to instead analyse smaller corpora, as similar as possible to the one used here, so as to at least partially estimate the impact of the size factor. Comparing the results reported here with those obtained by analysing a corpus of academic writing would also allows to determine their generalisability as conversation and academic writing strongly differ by their number of frequent n-grams (Biber, 1999). 

\nocite{AGR92,BER73,BES14,BES2018,BIB09,BIE13,CHU90,CHU91,COH60,DAI94,EK05,ELL15,EVE09,FRA00,GRI05,GRI10,HOL79,JON74,MOO04,NER10,OKE07,PEC10,PED96,PKB96,POW12,RAM15,STE03,WER06,ZEL95}

\section*{Acknowledgments}
The author is Research Associate of the Fonds de la Recherche Scientifique - FNRS (F\'ed\'eration Wallonie Bruxelles de Belgique).

\bibliography{acl2016}
\bibliographystyle{eacl2017}

\end{document}